%% file: main.tex
\documentclass[10pt,twocolumn,letterpaper]{article}

\usepackage[pagenumbers]{cvpr} % To force page numbers, e.g. for an arXiv version

\input{preamble}

\definecolor{cvprblue}{rgb}{0.21,0.49,0.74}
\usepackage[pagebackref,breaklinks,colorlinks,allcolors=cvprblue]{hyperref}

\def\paperID{43004} % *** Enter the Paper ID here
\def\confName{CVPR}
\def\confYear{2026}

\title{HERO: Hierarchical Embedding-Refinement for Open-Vocabulary Temporal Sentence Grounding in Videos}

\author{
Tingting Han$^{1}$ \quad
Xinsong Tao$^{1}$ \quad
Yufei Yin$^{1}$\quad
Min Tan$^{2}$\thanks{Corresponding author} \quad
Sicheng Zhao$^{3}$ \quad
Zhou Yu$^{2}$ \\
\small $^{1}$ Laboratory of Complex Systems
Modeling and Simulation, School of Computer
Science and Technology, Hangzhou Dianzi University \\
\small $^{2}$ Zhejiang Key Laboratory of Space
Information Sensing and
Transmission, Hangzhou Dianzi University \\
\small $^{3}$ Department of Psychological and Cognitive Sciences, Tsinghua University \\
{\tt\small 
ttinghan@hdu.edu.cn,
ttad@hdu.edu.cn,
yinyf@hdu.edu.cn}\\
{\tt\small
tanmin@hdu.edu.cn,
schzhao@tsinghua.edu.cn,
yuz@hdu.edu.cn
}
}

\begin{document}
\maketitle

\begin{abstract}
Temporal Sentence Grounding in Videos (TSGV) aims to temporally localize segments of a video that correspond to a given natural language query. Despite recent progress, most existing TSGV approaches operate under closed-vocabulary settings, limiting their ability to generalize to real-world queries involving novel or diverse linguistic expressions. To bridge this critical gap, we introduce the Open-Vocabulary TSGV (OV-TSGV) task and construct the first dedicated benchmarks—Charades-OV and ActivityNet-OV—that simulate realistic vocabulary shifts and paraphrastic variations. These benchmarks facilitate systematic evaluation of model generalization beyond seen training concepts. To tackle OV-TSGV, we propose \textbf{HERO}\footnote{Code: https://github.com/TTingHan-HDU/HERO} (\textbf{H}ierarchical \textbf{E}mbedding-\textbf{R}efinement for \textbf{O}pen-Vocabulary grounding), a unified framework that leverages hierarchical linguistic embeddings and performs parallel cross-modal refinement. HERO jointly models multi-level semantics and enhances video-language alignment via semantic-guided visual filtering and contrastive masked text refinement.
Extensive experiments on both standard and open vocabulary benchmarks demonstrate that HERO consistently surpasses state-of-the-art methods, particularly under open-vocabulary scenarios, validating its strong generalization capability and underscoring the significance of OV-TSGV as a new research direction.
\end{abstract}
\begin{figure}[tbp]
    \centering
    \includegraphics[width=0.95\linewidth]{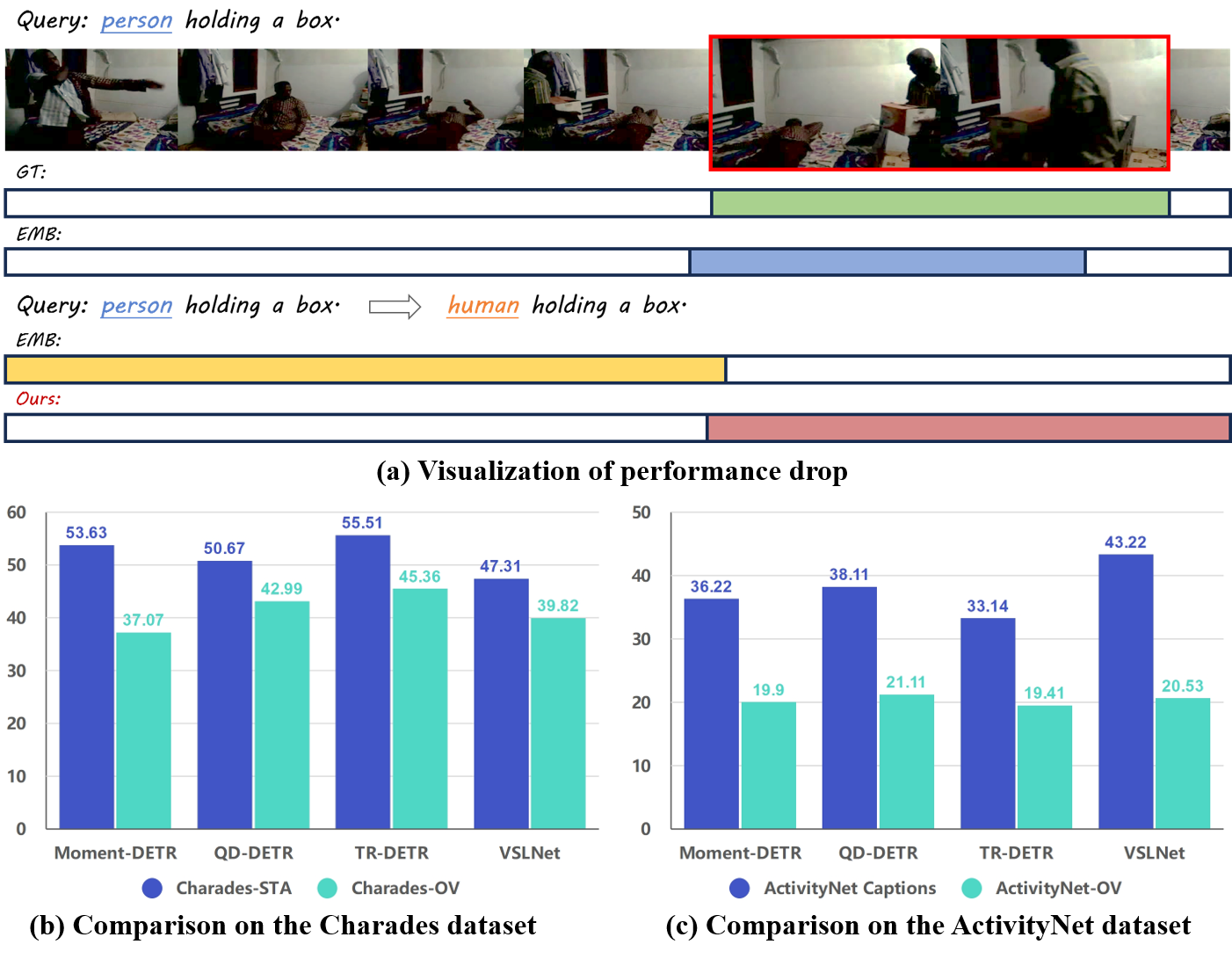}
    \caption{(a) Grounding visualization under open-vocabulary settings. (b) (c) Performance comparison on standard benchmarks (Charades-STA and ActivityNet Captions) and the proposed open-vocabulary datasets (Charades-OV, ActivityNet-OV).} 
    \label{fig:fig1}
    \vspace{-0.5cm}
\end{figure}
\section{Introduction}
Temporal sentence grounding in videos (TSGV) aims to localize a segment within an untrimmed video that corresponds to a given natural language query. This task plays a critical role in various video understanding applications such as content-based video retrieval, human-computer interaction, and intelligent surveillance. By bridging vision and language modalities, TSGV serves as a cornerstone for fine-grained cross-modal reasoning in temporal domains.

While Temporal Sentence Grounding in Video (TSGV) has made substantial progress, its development remains fundamentally limited by the constraints of existing datasets. Prior studies~\cite{lanCloserLookDebiased2023} reveal that TSGV models often overfit to dataset-specific biases—such as segment position and duration distributions—rather than learning robust video-language alignment. This is largely due to the high cost of precise temporal annotation, which restricts dataset scale and diversity. To mitigate such biases, recent works have explored strategies including causal intervention~\cite{yang2021deconfoundeddebias}, video shuffling~\cite{hao2022can}, curriculum augmentation~\cite{lan2023curriculum}, and theory-guided semantic learning~\cite{liu2022fewdebias}. However, a critical limitation persists: these methods are developed and evaluated solely under closed-vocabulary settings, where test queries closely resemble those seen during training. Consequently, they struggle to generalize to more realistic open-vocabulary scenarios involving novel or rare linguistic expressions (e.g., unseen objects, actions, or paraphrases).

% \begin{figure}[tbp]
%     \centering
%     \includegraphics[width=0.95\linewidth]{figure/figure1}
%     \caption{(a) Grounding visualization under open-vocabulary settings. (b) (c) Performance comparison on standard benchmarks (Charades-STA and ActivityNet Captions) and the proposed open-vocabulary datasets (Charades-OV, ActivityNet-OV).} 
%     \label{fig:fig1}
% \end{figure}

The vulnerability of existing methods to vocabulary shifts is clearly illustrated in \cref{fig:fig1}. As shown in (a), replacing a common token—e.g., “person” with the semantically equivalent but unseen term “human” in “person holds a box”—leads to significant localization failure in strong baselines like EMB~\cite{huang2022videoEMB}. Subfigures (b) and (c) further present aggregated results on both standard (Charades-STA, ActivityNet Captions) and open-vocabulary benchmarks (Charades-OV, ActivityNet-OV), revealing consistent and substantial performance degradation. These results highlight the brittleness of current closed-vocabulary approaches and their limited robustness in real-world, linguistically diverse scenarios.

To bridge this gap, we introduce the novel task of Open-Vocabulary Temporal Sentence Grounding in Video (OV-TSGV). In this setting, models are trained on queries composed of known concepts but must localize segments based on test-time queries containing novel objects, actions, or paraphrased expressions never seen during training. This formulation provides a principled benchmark to assess a model’s ability to generalize based on semantic abstraction and cross-modal compositionality—rather than memorization of training patterns.

To tackle the challenges of OV-TSGV, we contribute along two key dimensions. \textbf{(1) We construct the first benchmarks specifically designed for Open-Vocabulary TSGV}, simulating realistic vocabulary shifts and paraphrastic variations. These benchmarks provide a challenging and practical testbed for evaluating robust video-language grounding in open-world scenarios. \textbf{(2) We introduce HERO (Hierarchical Embedding and Refinement for Open-vocabulary grounding)}, a unified parallel framework that overcomes the limitations of closed-vocabulary models. By integrating hierarchical semantic representations with parallel cross-modal refinement, HERO significantly enhances alignment and generalization to unseen linguistic concepts and expressions.

Specifically, HERO adopts a hierarchical embedding and parallel processing architecture, comprising two key components: a Hierarchical Embedding Module (HEM) and a Cross-modal Filtering and Refinement Engine (CFRE). HEM extracts linguistic features across multiple semantic levels, enhancing the model’s ability to capture diverse linguistic patterns. Built upon these representations, CFRE operates in parallel across levels, applying two complementary submodules to refine fused video-text features: the Semantic-Guided Visual Filter (SGVF), which suppresses irrelevant visual content using textual cues for more precise grounding; and the Contrastive Masked Text Refiner (CMTR), which employs token masking and contrastive learning to improve textual robustness and generalization. This concurrent semantic encoding and refinement strategy enables HERO to achieve fine-grained, semantically aligned video-language grounding, effectively mitigating vocabulary brittleness in open-world scenarios.

Our main contributions are as follows:
\begin{itemize}
\item We introduce the first OV-TSGV benchmarks, Charades-OV and ActivityNet-OV, simulating realistic vocabulary shifts and paraphrastic variations for evaluating generalization in video-language grounding.
\item We propose HERO, a unified framework that combines hierarchical embedding and parallel cross-modal refinement to achieve robust alignment between visual content and diverse textual expressions.
\item We achieve state-of-the-art performance on both standard and open-vocabulary TSGV benchmarks, demonstrating the effectiveness of HERO and the importance of evaluating under open-vocabulary conditions.
\end{itemize}

\section{Related Work}
\subsection{Temporal Sentence Grounding in Videos}

TSGV methods are broadly categorized into proposal-based~\cite{xu2019multilevelQSPN, liu2018cross_based, jiang2019cross_based, wang2020temporally_based, HAN2025111621} and proposal-free~\cite{yuan2019findABLR, huang2022videoEMB, chen2021end_free, sun2023video_free} approaches. The former generate and rank candidate segments, while the latter directly predict frame-level boundaries, requiring precise cross-modal alignment.
Recent DETR-style models~\cite{lei2021detectingMoment-DETR, sun_zhou2024tr-detr, um2025watchk-detr} enhance proposal-free grounding via transformer-based matching and auxiliary tasks, yet remain limited to closed-vocabulary settings.
To address generalization under distribution shifts, debiasing techniques have emerged following the release of Charades-CD and ActivityNet-CD~\cite{lanCloserLookDebiased2023}, including causal intervention~\cite{yang2021deconfoundeddebias}, data augmentation~\cite{liu2022fewdebias}, contrastive learning~\cite{hao2022can, 10555372CoDebias}, and bias-guided training~\cite{clark2019don_modeldebias, han2021greedy_modeldebias, he2019unlearn_modeldebias, cadene2019rubi_modeldebias, nam2020learning_modeldebias}.

However, these methods assume shared vocabulary between training and testing. In contrast, we target the open-vocabulary setting, where test queries contain unseen concepts—addressing a critical gap in TSGV research.

\subsection{Open Vocabulary Learning}%feng2022promptdet,VLDet,gu2021open,
Open-vocabulary understanding aims to recognize novel classes not seen during training. Existing open-world learning methods have been extensively developed across various domains~\cite{ghiasi2022scaling,wu2023betrayed,ding2023maskclip,liu2023grounding, liu2025unified, 10130328}, leveraging techniques such as knowledge distillation~\cite{chen2023ovarnet}, prompt-based modeling~\cite{du2022learning, wu2023cora}, region-text pre-training~\cite{zareian2021open}, and region-text alignment~\cite{zang2022open}. However, despite these advances, open-vocabulary learning remains largely unexplored in the context of Temporal Sentence Grounding in Videos (TSGV).

Distinct from conventional open-vocabulary tasks, the concept of a ``class'' in TSGV is inherently ambiguous. We therefore define a class as any lexical unit within the query, treating each query as a set of classes. For instance, the query “person hold a box” corresponds to the class set \{\textit{person}, \textit{hold}, \textit{box}\}. Under this formulation, the Open-Vocabulary TSGV task requires the model to accurately localize the temporal segment corresponding to queries containing at least one unseen class during training.
\begin{figure}[tbp]
    \centering
    \includegraphics[width=1\linewidth]{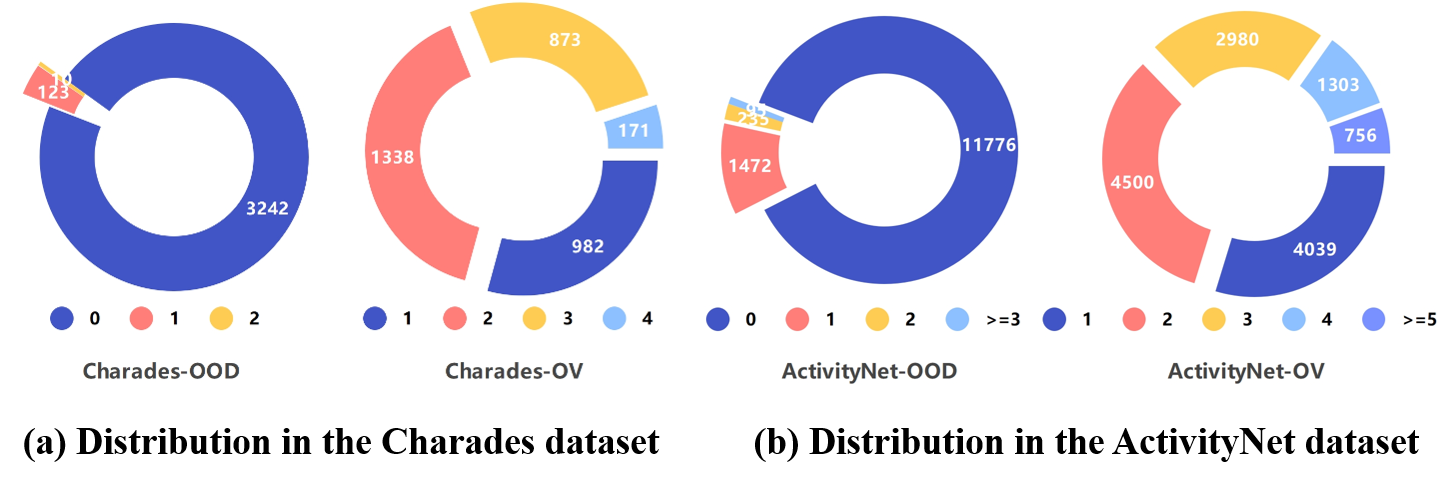}
    \caption{Distribution of sentences in the test set based on the number of words not present in the training vocabulary.}
    \label{fig:new_word}
\end{figure}
\begin{figure}[tbp]
    \centering
    \includegraphics[width=1\linewidth]{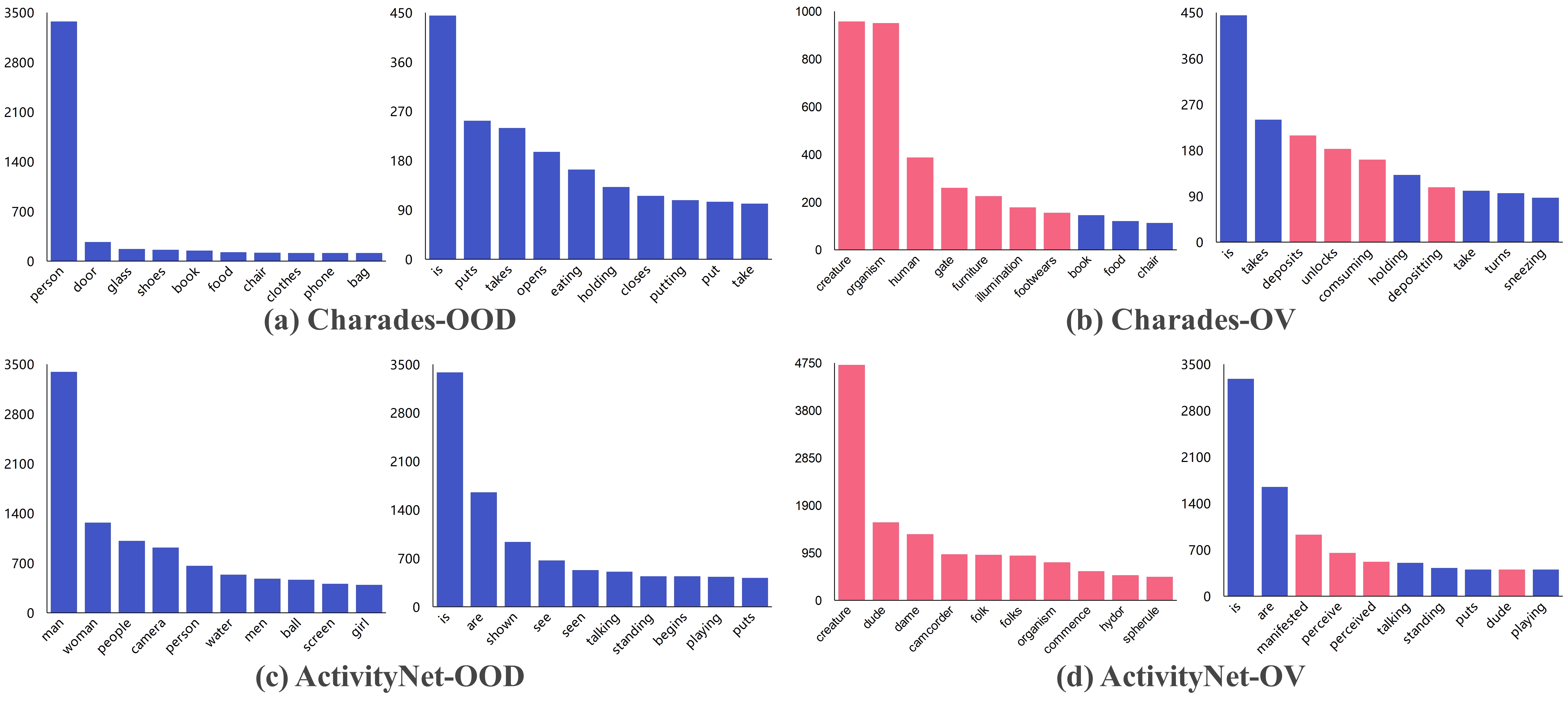}
    \caption{The distribution of words in the top ten frequent terms of the test set that did not appear in the training set. (a) test-ood in charades-cd. (b) test-ov in charades-ov. (c) test-ood in activitynet-cd. (d) test-ov in activitynet-ov. Red indicates that the word did not appear in the training set.}
    \label{new_word}
\end{figure}
\section{Proposed Method}
\subsection{Problem Definition}
\label{PD}
Unlike conventional open-vocabulary tasks where ``classes'' are predefined categories (e.g., object labels), the notion of a class in TSGV is inherently ambiguous due to the free-form nature of language queries. To address this, we define each lexical unit in a query as an individual class:
\begin{equation}
\qquad \mathcal{C}(q) \triangleq \{\, w \mid w \text{ is a lexical unit in query } q \,\}.    
\end{equation}
Here, $\mathcal{C}(q)$ denotes the set of lexical units (e.g., words or tokens) contained in the natural language query $q$, which we treat as semantic classes under the open-vocabulary grounding setting. 
For example, given the query:
\begin{equation}
q = \text{``person hold a box''} \Rightarrow \mathcal{C}(q) = \{\text{person},\,\text{hold},\,\text{box}\}.    
\end{equation}
Under this formulation, the Open-Vocabulary Temporal Sentence Grounding in Video (OV-TSGV) task is defined as follows: Given a video and a query $q$ such that $\exists c \in \mathcal{C}(q)$ with $c \notin \mathcal{C}_{\text{train}}$—i.e., at least one word in $q$ has not been observed during training—the model must accurately predict the corresponding temporal segment $(s, e)$ in the video that aligns with the query semantics.

\subsection{OV-TSGV Benchmark}

While prior TSGV research has primarily relied on Charades-STA (Gao et al., 2017) and ActivityNet Captions (Krishna et al., 2017), recent work has introduced Charades-CD and ActivityNet-CD (Lan et al., 2023b) to reduce dataset biases and better evaluate generalization. However, these benchmarks do not explicitly address the open-vocabulary setting, where test queries may involve novel or unseen concepts absent from the training data.
To address this limitation, we build upon Charades-CD and ActivityNet-CD to construct Charades-OV and ActivityNet-OV, where all textual queries in the \textit{test-ood} splits are rewritten to include at least one novel concept. The rewriting is performed using a large language model, followed by manual verification to ensure semantic coherence and vocabulary novelty. Notably, we retain the original train/val/test splits and sample counts from the corresponding CD datasets to ensure consistency and comparability. The resulting OV benchmarks enable a more focused and rigorous evaluation of model generalization under open-vocabulary temporal sentence grounding.

We further assess the lexical novelty of our proposed OV datasets by comparing them with existing distribution-shift benchmarks. Specifically, we measure the proportion of test sentences containing words unseen during training. As shown in \cref{fig:new_word}, 96.06\% of test sentences in the \textit{test-ood} split of Charades-CD and 86.73\% in ActivityNet-CD consist entirely of words present in the training vocabulary. These results indicate that, despite their goal of modeling distribution shifts, both benchmarks largely remain within a closed-vocabulary regime.
In contrast, our OV datasets explicitly enforce vocabulary divergence. Every test query in Charades-OV contains at least one unseen class—982 queries include one novel class, 1,338 include two, and 1,044 include three or more. Similarly, all test queries in ActivityNet-OV involve unseen concepts, with 4,039, 4,500, 2,980, and 2,059 queries containing one, two, three, and four or more novel classes, respectively.

As illustrated in \cref{new_word}, 
% the top-10 most frequent nouns and verbs in Charades-CD and ActivityNet-CD test splits all appear in their training sets, whereas our revised OV splits introduce high-frequency unseen terms—such as “human” in Charades-OV and “folk” in ActivityNet-OV—highlighting the enhanced lexical diversity and novelty of our benchmarks.
 highlights the top-10 most frequent nouns and verbs in the test-OOD splits of Charades-CD and ActivityNet-CD—all of which appear in their respective training sets. In contrast, our revised test-OV splits introduce high-frequency terms—such as "human" in Charades-OV and "folk" in ActivityNet-OV—that are entirely absent from the training vocabulary, further demonstrating the lexical diversity and novelty introduced by our OV benchmarks.
 \begin{figure*}[t]
    \centering
    \includegraphics[width=0.8\linewidth]{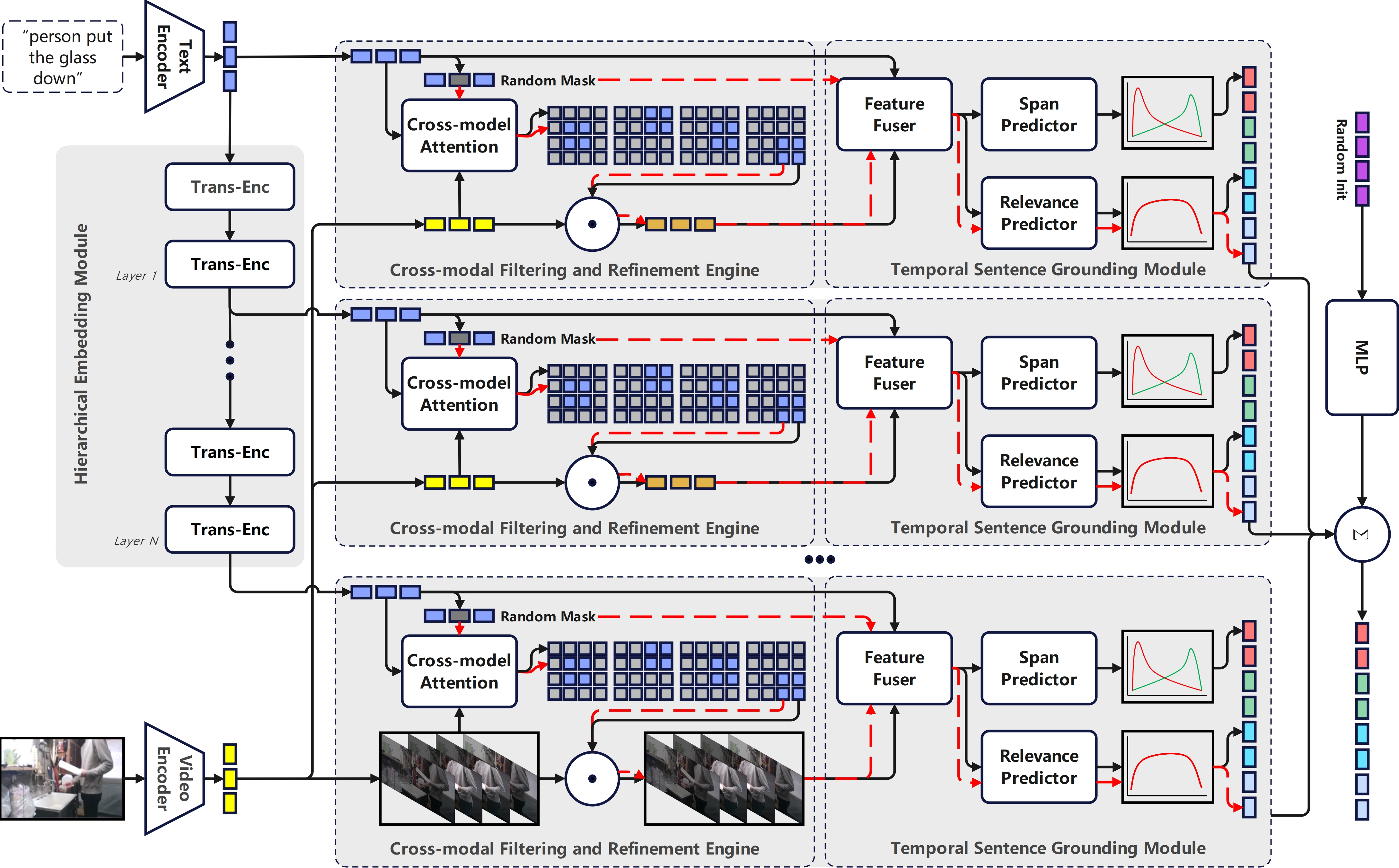}
    \caption{Framework overview of HERO. The Hierarchical Embedding Module (HEM) first extracts multi-level text representations from input queries. These hierarchical features then undergo parallel processing through Cross-modal Filtering and Refinement Engine (CFRE), where: (1) Semantic-Guided Visual Filters suppress irrelevant video content, while (2) Contrastive Masked Text Refiners enhance linguistic robustness. Finally, the refined cross-modal features from each CFRE branch are fed into a Temporal Grounding Module to produce hierarchical predictions, which are aggregated via weighted summation for the final temporal localization result.}
    \label{fig:over_view}
\end{figure*}
For completeness, Charades-OV comprises 4,564 videos with 11,071 queries for training, 333 videos with 859 queries for validation, and 1,442 videos with 3,364 queries for testing. ActivityNet-OV contains 10,984/746/2,450 videos and 51,415/3,521/13,578 queries for the training/validation/test splits, respectively.

\subsection{HERO Network for OV-TSGV}

\Cref{fig:over_view} depicts the overall architecture of our proposed framework, \textbf{HERO} (Hierarchical Embedding-Refinement for Open-vocabulary grounding), which employs a hierarchical embedding and parallel processing strategy. Given an input video and a textual query, we first extract visual features \( V = \{v_{t}\}_{t=1}^{T} \) and textual features \( Q = \{q_{i}\}_{i=1}^{L} \) via dedicated encoders. To capture multi-granularity linguistic semantics, the Hierarchical Embedding Module (HEM) transforms the query into a set of representations \( \{Q_i\}_{i=1}^{N} \) spanning from low-level lexical tokens to high-level semantic abstractions. The hierarchical features are processed in parallel through the Cross-modal Filtering and Refinement Engine (CFRE). Within each branch, CFRE enhances video-language alignment by generating refined representations. This process includes a Semantic-Guided Visual Filter (SGVF), which suppresses irrelevant visual information guided by textual cues, and a Contrastive Masked Text Refiner (CMTR), which strengthens textual robustness via masked token prediction and contrastive learning. The refined features are then passed to the Temporal Sentence Grounding module, composed of two prediction heads: one estimates temporal boundaries \( (P^s_i, P_i^e) \), and the other computes alignment scores \( \{RS_i, RS^m_i\} \). Each level produces an output tuple \( \textit{O}_i = \{(P_i^s, P_i^e), RS_i, RS^m_i\} \), and a learnable weighted aggregation of these outputs yields the final grounding prediction \( (s, e) \). 
This design facilitates precise and semantically consistent temporal sentence grounding under open-vocabulary settings.

\subsubsection{Hierarchical Embedding Module}

Open-vocabulary temporal grounding requires models to handle queries with rare or unseen expressions. Token-level encodings often fail to capture semantic equivalence across varied phrasings (e.g., ``boy grabs skateboard'' vs. ``kid picks up object''), limiting generalization. To address this, we introduce hierarchical semantic abstraction over text features, enabling multi-level representations from lexical to conceptual granularity.

We implement a Transformer-based Hierarchical Embedding Module (HEM) using a 6-layer encoder, extracting outputs from layers 2, 4, and 6, along with the input embedding. This yields four levels of semantic representations that enhance robustness to linguistic variation.
\begin{equation}
Q_{0} = Q,
\end{equation}
\begin{equation}
Q_{1} = \text{TransformerEncoder}_{2}(Q_{0}),
\end{equation}
\begin{equation}
Q_{i} = \text{TransformerEncoder}_{2i}(Q_{i-1}), \text{for } i = 2, 3,
\end{equation}
where, \(Q_{0}\) denotes the initial token-level text features, and \(Q_{i}\) for \(i\geq 1\) represents progressively abstracted representations derived from deeper layers of the Transformer Encoder. This hierarchical modeling strategy enables our framework to encode both low-level lexical cues and high-level semantic concepts, which is essential for robust video-text alignment under open-vocabulary settings.

\subsubsection{Cross-modal Filtering and Refinement Engine}

In open-vocabulary temporal sentence grounding, effective alignment between video and text modalities is challenged by both noisy or irrelevant visual information and diverse, potentially unseen linguistic expressions. To address these issues, we design the Cross-modal Filtering and Refinement Engine (CFRE), which jointly refines multimodal features by suppressing irrelevant visual cues with a Semantic-Guided Visual Filter and enhancing the robustness of textual embeddings by the Contrastive Masked Text Refiner. This dual refinement mechanism is critical for improving cross-modal interaction and achieving reliable grounding under open-vocabulary conditions. 

\subsubsection{Semantic-Guided Visual Filter}
To suppress irrelevant visual content and highlight query-relevant regions, we design a semantic-guided visual filter based on a cross-attention mechanism. Given video features \( V \) and text features \( Q_{i}, i \in \{0,1,2,3\} \) of a specific level, we compute attention weights using \( V \) as the query and \( Q_{i} \) as the key and value. The resulting attention weights are passed through a sigmoid activation to generate soft relevance coefficients, which modulate the original video features via element-wise multiplication, thereby suppressing background noise and enhancing relevant visual content.
\begin{equation}
V_{i}^{\text{attn}} = \text{Softmax}\left(\frac{V Q_{i}^T}{\sqrt{d_{k}}}\right) Q_{i},
\end{equation}
\begin{equation}
\hat{V}_i = V \odot \text{Sigmoid}(V_{i}^{\text{attn}}),
\end{equation}
where \( d_k \) is the key dimensionality and \( \odot \) denotes element-wise multiplication. The softmax function normalizes the attention scores, while the sigmoid function ensures the relevance coefficients are bounded between 0 and 1. $\hat{V}_i$ represents the enhanced visual representation under the guidance of textual semantics. This operation enhances the alignment between modalities by adaptively filtering out background noise and amplifying semantically relevant visual signals.

\subsubsection{Contrastive Masked Text Refiner}
To improve the robustness and semantic generalization of textual representations, we adopt a contrastive learning strategy that encourages the model to maintain consistent cross-modal relevance even when parts of the textual query are missing or perturbed. Specifically, we randomly mask a subset of tokens from the original query $Q_{i}$ to generate a corrupted variant $Q_{i}^m$, formulated as:
\begin{equation}
    Q_i^m=\text{RandomMask}(Q_i).
\end{equation}

The masked query $Q_i^m$ is also fed into the \textbf{Semantic-Guided Visual Filter} along with the video features $V$
, producing a relevance-enhanced visual representation $\hat{V}_i^{m}$. Both the original pair $\{Q_i,\hat{V}_i\}$ and the corrupted pair $\{Q_i^m, \hat{V}_i^{m}\}$ are then processed by the \textbf{Temporal Sentence Grounding Module}, which computes their respective relevance scores $RS_{i}$ and $RS_i^m$, respectively. The relevance scores $RS_i$ and $RS_i^m$ from different layers are aggregated via a learnable weighted summation to obtain unified scores $RS$ and $RS^m$. We then compute the KL divergence between them as the training loss:
\begin{equation}
\mathcal{L}_{\text{CL}} = D_{\text{KL}}(RS \| RS^m ).    
\end{equation}

By minimizing this discrepancy, the model is trained to promote consistency in video-text alignment despite input perturbations, which enhances its robustness against noisy or incomplete textual inputs.

\subsubsection{Temproal Sentence Grounding Module}
To ensure broad applicability, our framework is designed as a plug-and-play enhancement module that can be seamlessly integrated into various temporal sentence grounding (TSG) architectures. While existing TSG models commonly adopt a two-stage paradigm—consisting of a Feature Fuser and a Span Predictor—our method imposes no restrictions on their specific implementations. Instead, it functions independently as an auxiliary component, enhancing the model’s robustness and generalization, especially in open-vocabulary scenarios.

In our experiments, we instantiate our framework on top of the EMB model~\cite{huang2022videoEMB}, demonstrating its compatibility and effectiveness. Specifically, we first extract fused features from the visual-textual inputs using a Feature Fuser. Given the original pair $\{Q_i,\hat{V}_i\}$ and the corrupted pair $\{Q_i^m, \hat{V}_i^{m}\}$, the fused features are computed as:
\begin{equation}
    F_i = \text{FeatureFuser}(\hat{V}_i, Q_i),
\end{equation}
\begin{equation}
    F^m_i = \text{FeatureFuser}(\hat{V}_i^m, Q_i^m).
\end{equation}

Subsequently, the Span Predictor estimates the temporal boundaries $(P_i^s,P_i^e)$ from the original fused feature $F$:
\begin{equation}
    (P_i^s,P_i^e) = \text{SpanPredictor}(F_i).
\end{equation}

To measure cross-modal relevance, we employ a MLP as the Relevance Predictor, producing the relevance scores corresponding to the fused features $F_i$ and $F^m_i$:
\begin{equation}
    RS_i = \text{MLP}(F_i),
\end{equation}
\begin{equation}
    RS^m_i = \text{MLP}(F^m_i).
\end{equation}

Thus, the output of the TSG module comprises the temporal span predictions and the relevance scores, denoted as $O_{i}=\{(P_i^s,P_i^e),RS_i,RS^m_i\}.$

\begin{figure*}[tbp]
    \centering
    \includegraphics[width=0.98\linewidth]{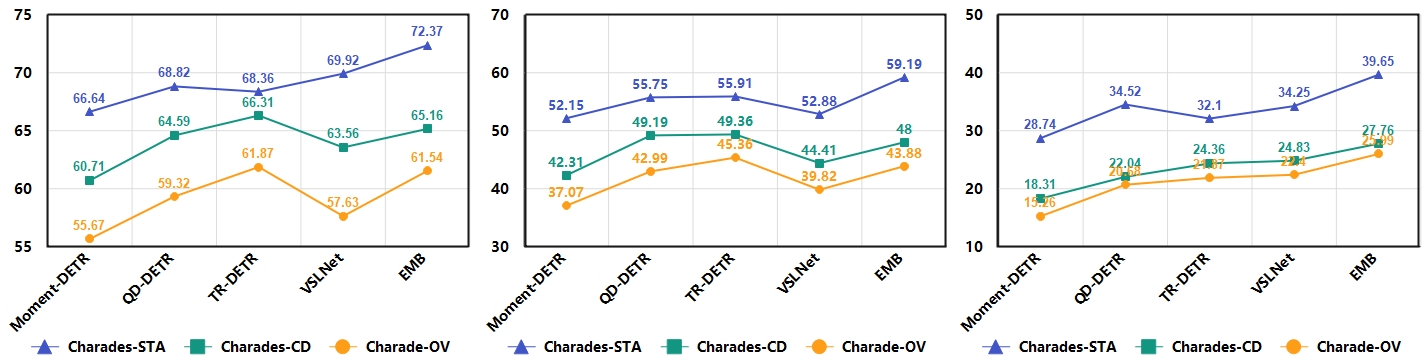}
\caption{Performance comparison of state-of-the-art methods on different versions of the Charades dataset. The x-axis denotes different TSG models, and each line corresponds to a specific dataset variant. From left to right, the sub-figures report results for R1@0.1, R1@0.5, and R1@0.7, respectively.}
    \label{dataset_cmp}
    %\vspace{-0.3cm}
\end{figure*}

\subsubsection{Output and Loss Functions}
Based on \(N\) parallel branches, each capturing the model’s understanding of the input at a different level of abstraction, we obtain \(N\) intermediate outputs. To derive the final prediction, we employ a learnable weighted aggregation mechanism that adaptively fuses these outputs. Specifically, a set of scalar weights \(\{W_i\}_{i=1}^N\) is learned by feeding randomly initialized vectors \(T_i \in \mathbb{R}^{d}\) into a lightweight MLP. The final output is then computed as a weighted summation of the branch outputs:
\begin{equation}
    \textit{O} = \sum_{i=1}^N W_{i} \textit{O}_{i},
\end{equation}
\begin{equation}
    W_{i} = \text{MLP}(T_{i}) , \quad W_i \in \mathbb{R},
\end{equation}
\begin{equation}
    T_{i} = \text{RandomInit}, \quad T_i \in \mathbb{R}^{d},
\end{equation}
where \(\textit{O}\) is the final output of the model, \(\textit{O}_{i}\) is the output from the \(i\)-th level of abstraction, \(W_{i}\) is the learnable weight for the \(i\)-th level, and \(T_{i}\) is a randomly initialized tensor. The MLP is used to transform the randomly initialized tensor \(T_{i}\) into the weight \(W_{i}\), which then modulates the contribution of \(\textit{O}_{i}\) to the final \(\textit{O}\).

\textbf{Final Loss.} 
The total training loss combines three components in a weighted manner, each supervising a different aspect of model learning:
\begin{equation}
\mathcal{L} = \mathcal{L}_{\text{TSGV}} + \lambda_1 \mathcal{L}_{\text{RS}} + \lambda_2 \mathcal{L}_{\text{CL}},
\end{equation}
where \(\mathcal{L}_{\text{TSGV}}\) denotes the primary loss for the TSGV task~\cite{huang2022videoEMB}. $\mathcal{L}_{\text{RS}}$ is the relevance score loss, which constrains the predicted relevance distributions to align with the ground-truth temporal supervision over video frames. $\mathcal{L}_{\text{CL}}$ is the contrastive learning loss introduced in the \textbf{Contrastive Masked Text Refiner} module, designed to improve robustness against input perturbations. The hyperparameters \(\lambda_1\) and \(\lambda_2\) are tunable coefficients that control the relative importance of the relevance score loss and the contrastive learning loss, respectively.

\textit{Relevance Score Loss.}  
To guide the learning of frame-level relevance predictions, we define a relevance score loss \(\mathcal{L}_{\text{RS}}\), which averages the binary cross-entropy (BCE) losses computed from both the original and the perturbed (e.g., masked) textual inputs:
\begin{equation}
\mathcal{L}_{\text{RS}} = \frac{1}{2} \left( \mathcal{L}_{\mathrm{BCE}}(\boldsymbol{V}, \boldsymbol{RS}) + \mathcal{L}_{\mathrm{BCE}}(\boldsymbol{V}, \boldsymbol{RS}^m) \right),
\end{equation}
where \(\boldsymbol{RS}\) and \(\boldsymbol{RS}^m\) denote the relevance scores predicted from the original query and the masked version of the query, respectively.
The BCE loss measures the alignment between predicted relevance scores and ground-truth frame-level annotations. Specifically, it is formulated as:
\begin{align}
\mathcal{L}_{\mathrm{BCE}}(\boldsymbol{V}, \boldsymbol{RS}) = -\sum_{\boldsymbol{v}_t} \Big[\, &p(\boldsymbol{v}_t) \log(RS(\boldsymbol{v}_t)) \notag \\
&+ (1 - p(\boldsymbol{v}_t)) \log(1 - RS(\boldsymbol{v}_t)) \,\Big],
\end{align}
where \(p(\boldsymbol{v}_t)\) is the binary ground-truth signal indicating whether frame \(\boldsymbol{v}_t\) is relevant to the query, and \(RS(\boldsymbol{v}_t)\) is the model's predicted relevance score for that frame.

\begin{table*}[t]
    \renewcommand{\tabcolsep}{3.3mm}
    \centering
    \caption{Temporal Sentence Grounding performance on the \textit{test-ov} splits of Charades-OV and ActivityNet-OV. \textbf{Bold} and \underline{underlined} numbers denote the best and second-best results, respectively.}
        \begin{tabular}{ccccccc}
            \Xhline{3\arrayrulewidth}
            \rule{0pt}{10.0pt} \multirow{2}{*}[-0.35em]{\bf Method} & \multicolumn{3}{c}{\textbf{Charades-OV}} & \multicolumn{3}{c}{\textbf{Activitynet-OV}}\\ 
            \cmidrule(lr){2-4} \cmidrule(l){5-7}
            & R1@0.3 & R1@0.5 & R1@0.7 & R1@0.3 & R1@0.5 & R1@0.7 \\
            \midrule
            Moment-DETR~\cite{lei2021detectingMoment-DETR} & 55.67 & 37.07 & 15.26 & 34.26 & 19.90 & 9.16 \\
            VSLNet~\cite{zhang2021naturalVSLNet}  &  57.63   & 39.82 & 22.40 & 38.64 & 20.53 & 10.19  \\
            EMB~\cite{huang2022videoEMB}      &  61.54  & 43.88 & \underline{25.99} & \underline{40.22} & \underline{21.70} & \underline{10.78} \\
            QD-DETR~\cite{moon2023queryqd-detr}   & 59.32  & 42.99 & 20.68 & 35.71 & 21.11 & 9.98\\
            TR-DETR~\cite{sun_zhou2024tr-detr}    & \underline{61.87} & \underline{45.36} & 21.87 & 34.28 & 19.41 & 9.00 \\
            
            \hdashline\rule{0pt}{9.5pt}
            \textbf{HERO (Ours)}      & \textbf{64.74}  & \textbf{45.51} & \textbf{27.20} & \textbf{42.78} & \textbf{25.23} & \textbf{12.18} \\
            \Xhline{3\arrayrulewidth}
             
        \end{tabular}
    \label{ov-sets}
\end{table*}

\section{Experiments}
\subsection{Experimental Setup}

\textbf{Datasets and Evaluation Metrics.}
% We conduct experiments on the Charades-\{STA, OV\} and ActivityNet-OV datasets. 
%
% For evaluation, we adopt the metrics \textbf{R@n, IoU@\(m\)} and \textbf{mean IoU (mIoU)}. Following prior work, we set \(n=1\) and \(m \in \{0.3, 0.5, 0.7\}\) in our evaluations. For more details, please refer to the supplementary material.
We adopt the pre-trained I3D~\cite{carreira2017quoi3d} video features for Charades and ActivityNet as video encoding provided by our baseline model~\cite{huang2022videoEMB} and 300D GloVe~\cite{pennington2014glove} embeddings to encode the querys, respectively.
All hidden representations are set to a fixed dimensionality of 128. For the Transformer Encoder in HEM, we adopt a 6-layer architecture and extract intermediate outputs from every 2 layers.
The model is trained for 20 epochs with a batch size of 16. Optimization is performed using the Adam optimizer, with a linearly decaying learning rate initialized at 0.0005 and gradient clipping set to 1.0. The loss weight coefficients are empirically set to $\lambda_1 = \lambda_2 = 0.1$. %All experiments are conducted on a system equipped with an NVIDIA GeForce RTX 2080 Ti GPU and an Intel Xeon Gold 5218 CPU @ 2.30 GHz.

For evaluation, we adopt the commonly used metrics \textbf{R@n, IoU@\(m\)} and \textbf{mean IoU (mIoU)}. Specifically, R@n, IoU@\(m\) denotes the percentage of test samples where at least one of the top-\(n\) predicted segments has an Intersection-over-Union (IoU) with the ground-truth segment greater than threshold \(m\). The \textbf{mIoU} metric computes the average IoU between the top-1 prediction and the ground truth over all test samples. Following prior work, we set \(n=1\) and \(m \in \{0.3, 0.5, 0.7\}\) in our evaluations.
%------------------------------------ Table 3 - Charades-STA
%##################################################################################################
\begin{table}[tbp]
    \renewcommand{\tabcolsep}{1.6mm}
    \centering
    \caption{Experimental results on the Charades-STA dataset. \textbf{Bold}/\underline{underlined} fonts indicate the best/second-best results. }
		\begin{tabular}{ccc}
            \Xhline{3\arrayrulewidth}
            \rule{0pt}{10.0pt}
            \textbf{Method} & R1@0.5 & R1@0.7 \\
            \hline
            \rule{0pt}{9.5pt}
            CTRL~\cite{gao2017tallCTRL} & 23.63 & 8.89 \\
            MAN~\cite{DBLP:conf/cvpr/ZhangDWWD19MAN} & 46.53 & 22.72 \\
            VSLNet~\cite{zhang2021naturalVSLNet} & 47.31 & 30.19 \\
            EMB~\cite{huang2022videoEMB} & 58.33 & \underline{39.25} \\
            QD-DETR~\cite{moon2023queryqd-detr} & 50.67 & 31.02 \\
            TR-DETR~\cite{sun_zhou2024tr-detr}  & 55.51 & 33.66 \\
            TaskWeave~\cite{yang2024task} & 53.36 & 31.40\\
            BM-DETR~\cite{10943446Ba-Detr} & 54.42 & 33.84 \\
            LLMEPET~\cite{jiang2024prior} & - & 36.49 \\
            $R^2$-tuning~\cite{liu2024r}  & 59.78 & 37.02\\
            FlashVTG~\cite{cao2025flashvtg} & \underline{60.11} & 38.01 \\
            \cdashline{1-3}
            \rule{0pt}{9.5pt}
            \textbf{HERO (Ours)} & \textbf{61.05} & \textbf{41.29} \\
            \Xhline{3\arrayrulewidth}
            \end{tabular}

    \label{charades}
\end{table}
%################################################################################################## 

\subsection{Comparisons on Different Datasets}

To evaluate model generalization and highlight the challenges of open-vocabulary settings, we compare five state-of-the-art methods on three Charades variants: the original Charades-STA, the distribution-shifted Charades-CD, and our proposed Charades-OV.
As shown in \cref{dataset_cmp}, all models exhibit performance drops on Charades-CD and Charades-OV compared to Charades-STA, with the decline more pronounced on Charades-OV. Unlike Charades-CD, which involves distribution shifts, Charades-OV further introduces novel expressions and unseen concepts, presenting a greater challenge. These results underscore the need for TSG models with stronger semantic generalization and robustness to vocabulary variation in real-world scenarios.

\subsection{Comparison with SOTA Methods}
\subsubsection{Results on Open-vocabulary TSGV Datasets}
\Cref{ov-sets} presents a comprehensive comparison between our method and existing state-of-the-art approaches on the proposed open-vocabulary (OV) benchmarks. Our approach achieves consistent and significant improvements across all settings. Notably, it surpasses previous methods by 2.56\% at R1@0.3, 3.53\% at R1@0.5 and 1.40\% at R1@0.7 on ActivityNet-OV, and achieves an additional 1.21\% gain at R1@0.7 and 2.87\% at R1@0.3 on Charades-OV. These substantial improvements clearly demonstrate the effectiveness of our method in addressing the vocabulary shift inherent to OV scenarios. The results confirm that our framework better captures cross-modal relevance and exhibits stronger generalization to unseen concepts during inference.

\subsubsection{Result on Charades-STA datasets}

\Cref{charades} compares our method with prior state-of-the-art approaches on Charades-STA. Proposal-based models (e.g., CTRL~\cite{gao2017tallCTRL}, ACL~\cite{DBLP:conf/wacv/GeGCN19ACL}) show limited performance, while proposal-free methods such as VSLNet~\cite{zhang2021naturalVSLNet} and DETR-style models~\cite{moon2023queryqd-detr, sun_zhou2024tr-detr} perform better. EMB~\cite{huang2022videoEMB} achieves 58.33\% R1@0.5 and 39.25\% R1@0.7, while our method reaches 61.05\% and 41.29\%, setting a new state-of-the-art under closed-vocabulary settings.

\subsection{Ablation Study}

\subsubsection{Component-wise Ablation Analysis}
\Cref{ablation_study} reports the ablation results of the two main components in HERO: HEM and CFRE. 
Individually adding HEM or either submodule of CFRE consistently improves performance over the baseline. Notably, combining all components yields the best results across all metrics, demonstrating their complementary roles. These results validate the effectiveness of our design in enhancing temporal sentence grounding through hierarchical abstraction and refined cross-modal alignment.

%##################################################################################################
%------------------------------------ Table 4 - Module ablation
%##################################################################################################
\begin{table}[t]
    \renewcommand{\tabcolsep}{2.3mm}
    \centering
    \small
   \caption{Ablation results on Charades-OV \textit{test-ov}, showing the effect of HEM and CFRE, where CFRE consists of \textit{SGVF} and \textit{CMTR}. \textbf{Bold}/\underline{Underlined} indicate the best and second-best results, respectively.}
\begin{tabular}{cccccc}
            \Xhline{3\arrayrulewidth}
            \rule{0pt}{10.0pt}
            \multirow{2}{*}[-0.8ex]{\textbf{HEM}} & \multicolumn{2}{c}{\textbf{CFRE}} & \multirow{2}{*}[-0.8ex]{R1@0.5} & \multirow{2}{*}[-0.8ex]{R1@0.7}  & \multirow{2}{*}[-0.8ex]{mIoU}\\
            \cmidrule(lr){2-3}
            & \textit{SGVF} & \textit{CMTR} & & \\
            \midrule
            - & - & - & 42.31 & 25.22 & 41.93\\
            \cdashline{1-6}
            \rule{0pt}{10.0pt}
            \checkmark & - & - & 42.64 & 24.89 & 42.03\\
            - & \checkmark & - & 43.58 & 25.72 & 43.24\\
            - & - & \checkmark & 42.24 & 24.81 & 42.03\\
            \checkmark & \checkmark & - & \underline{45.10} & \underline{26.70} & \underline{44.00}\\
            \checkmark & - & \checkmark & 42.73 & 24.63 & 42.01\\
            - & \checkmark & \checkmark & 44.82 & 25.21 & 43.39\\
            \checkmark & \checkmark & \checkmark & \textbf{45.51} & \textbf{27.20} & \textbf{44.86}\\
            \Xhline{3\arrayrulewidth}
        \end{tabular}

    \label{ablation_study}
\end{table}

%------------------------------------ Figure - Num of inputs
%##################################################################################################
\begin{figure}[tbp]
    \centering
    \includegraphics[width=1\linewidth]{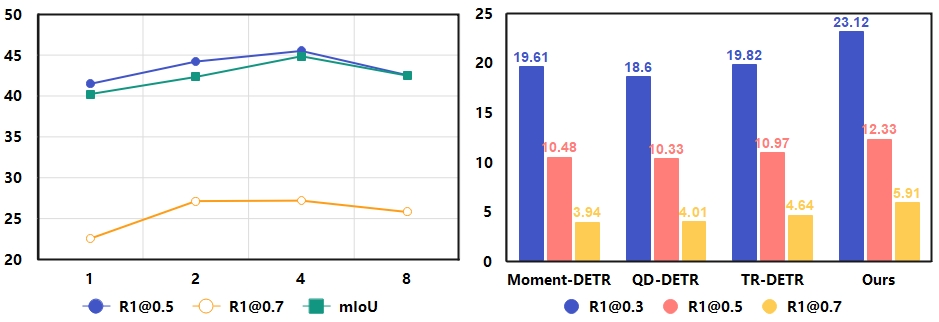}
    \caption{(Left) Performance comparison under different numbers of input pairs fed into the HEM module. (Right) Cross-dataset generalization results, where models trained on Charades-CD are evaluated on ActivityNet-CD.}
    \label{ablation_num}
    %\vspace{-0.3cm}
\end{figure}
%##################################################################################################

%------------------------------------ Figure - Visu 1
%##################################################################################################
\begin{figure}[tbp]
    \centering
    \includegraphics[width=0.99\linewidth]{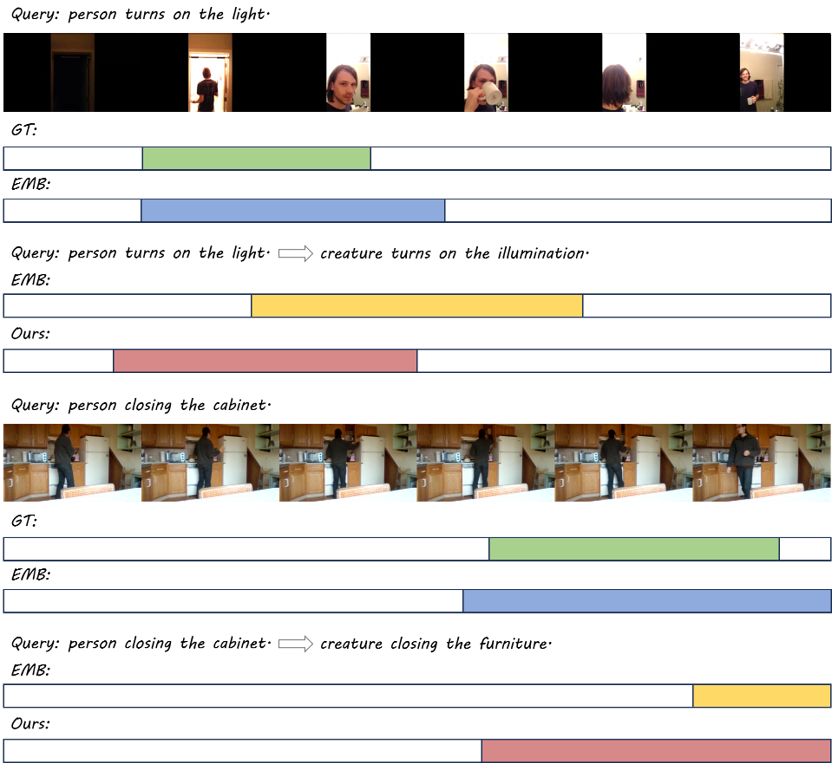}
    \caption{Qualitative comparison of per-sample grounding performance between our HERO and EMB~\cite{huang2022videoEMB} on the Charades-OV.}      \label{visu_1}
\end{figure}
%##################################################################################################

\subsubsection{Number of the Inputs in HEM}
\Cref{ablation_num} (Left) examines the impact of parallel depth in our HEM framework. The results demonstrate that employing four parallel layers achieves the optimal balance. With only two layers, the model tends to over-focus on superficial lexical and syntactic cues, placing excessive emphasis on literal details. Conversely, eight layers lead to over-abstraction, reducing sensitivity to fine-grained token information. The four-layer configuration strikes an effective balance: lower layers capture how the words are expressed, while upper layers encode what the words truly mean.

\subsubsection{Cross-dataset Generalization}
\Cref{ablation_num} (Right) presents the cross-dataset evaluation results, where models are trained on the Charades-CD training split and tested on the ActivityNet-CD test split. Our method achieves improvements of 3.3\% at R1@0.3, 1.36\% at R1@0.5 and 1.27\% at R1@0.7 compared to previous approaches, highlighting its stronger generalization ability across domains.

\subsection{Visualization Results}
\Cref{visu_1} presents a qualitative comparison between our HERO model and the baseline EMB~\cite{huang2022videoEMB} on representative samples from the Charades-OV dataset. Under open-vocabulary conditions, EMB exhibits notable grounding errors due to its limited ability to handle unseen expressions. In contrast, HERO maintains accurate localization across diverse queries, demonstrating its enhanced robustness and generalization to novel linguistic inputs.

\section{Conclusion}
In this paper, we study the underexplored yet practical task of Open-Vocabulary Temporal Sentence Grounding in Video (OV-TSGV), which requires localizing video segments for queries containing unseen or paraphrased expressions. To support this, we construct two benchmarks—Charades-OV and ActivityNet-OV—that simulate realistic vocabulary shifts. We also propose \textbf{HERO}, a hierarchical and parallel framework featuring (1) a Hierarchical Embedding Module (HEM) for multi-level linguistic semantics abstraction, and (2) a Cross-modal Filtering and Refinement Engine (CFRE) with Semantic-Guided Visual Filtering and Contrastive Masked Text Refinement to boost video-language interaction and robustness.
Experiments show HERO achieves sota results under both standard and OV settings. Future work will explore few-shot adaptation, continual learning, and broader multimodal grounding under open-world conditions.

\section*{Acknowledgments}
This work is partly supported by the Zhejiang Province Natural Science Foundation (Nos. LMS26F020018, LQN26F020053, LRG26F020001), the Beijing Natural Science Foundation (No. L252009), the National Natural Science Foundation of China (Nos. 62422204, 62472133, 62571294, 62441614), the Key Research and Development Program of Zhejiang Province (No. 2025C01026) and the Scientific Research Innovation Capability Support Project for Young Faculty.

{
    \small
    \bibliographystyle{ieeenat_fullname}
    \bibliography{main}
}


\end{document}

%% file: preamble.tex
\usepackage{algorithm}
\usepackage{algorithmic}

\usepackage{multirow, array, booktabs,makecell,xspace,bm}
\usepackage{arydshln}
\usepackage{amsfonts}
\usepackage{amssymb}

\usepackage{mathtools}
\usepackage{pifont}% http://ctan.org/pkg/pifont
\usepackage{dsfont}
\usepackage{lipsum}
\usepackage{xcolor}